\documentclass[authoryear,preprint,12pt]{elsarticle}

\usepackage{amssymb}

\usepackage{xcolor}
\usepackage{url}
\usepackage{graphicx,caption}
\usepackage{float}
\usepackage{textcomp, gensymb}
\usepackage{gensymb}
\usepackage{amsmath,amssymb}
\usepackage{verbatimbox}
\usepackage[outdir=./]{epstopdf}

\journal{Computers Environment and Urban Systems}

\begin{document}

\begin{frontmatter}



\title{DeepVerge: Classification of Roadside Verge Biodiversity and Conservation Potential}


\author[inst1]{Andrew Perrett}

\affiliation[inst1]{organization={School of Computer Science, University of Lincoln},
            addressline={Brayford Pool}, 
            city={Lincoln},
            postcode={LN6 7TS}, 
            state={Lincolnshire},
            country={UK}}

\author[inst4]{Charlie Barnes}
\author[inst2]{Mark Schofield}
\author[inst3]{Lan Qie}
\author[inst1]{Petra Bosilj}
\author[inst1]{James M. Brown}

\affiliation[inst2]{organization={Plantlife},
            addressline={Brewery House}, 
            city={Salisbury},
            postcode={SP1 2AP}, 
            state={Wiltshire},
            country={UK}}

\affiliation[inst3]{organization={School of Life Sciences, University of Lincoln},
            addressline={Brayford Pool}, 
            city={Lincoln},
            postcode={LN6 7TS}, 
            state={Lincolnshire},
            country={UK}}


\affiliation[inst4]{organization={Lincolnshire Wildlife Trust},
            addressline={Banovallum House}, 
            city={Horncastle},
            postcode={LN9 5HF}, 
            state={Lincolnshire},
            country={UK}}
            
\begin{abstract}
Open space grassland is being increasingly farmed or built upon, leading to a ramping up of conservation efforts targeting roadside verges. Approximately half of all UK grassland species can be found along the country's 500,000 km of roads, with some 91 species either threatened or near threatened. Careful management of these ``wildlife corridors'' is therefore essential to preventing species extinction and maintaining biodiversity with grassland habitats. Wildlife trusts have often enlisted the support of volunteers to survey roadside verges and identify new ``Local Wildlife Sites'' as areas of high conservation potential. Using volunteer survey data from 3,900 km of roadside verges alongside publicly available street-view imagery, we present \textit{DeepVerge}; a deep learning-based method that can automatically survey sections of roadside verge by detecting the presence of positive indicator species. Using images and ground truth survey data from the rural county of Lincolnshire, DeepVerge achieved a mean accuracy of 88\%. Such a method may be used by local authorities to identify new local wildlife sites, and aid management and environmental planning in line with legal and government policy obligations, saving thousands of hours of manual labour.
\end{abstract}



\begin{keyword}
Google Street View \sep Convolutional Neural Network \sep Machine Learning \sep Life on the Verge \sep Local Wildlife Sites \sep Remote Surveying \sep CNN \sep Lincolnshire Wildlife Trust

\end{keyword}

\end{frontmatter}


\section{Introduction and Motivation}
\label{sec:intro}
Over the past century, open space grassland and wildflower meadows have been increasingly farmed or built upon. At the same time, the number of roads (and therefore roadside verges) has increased, making the proportion of grassland within roadside verges much greater. 
The 2011 UK National Ecosystem Assessment \citep{UKNEA2011} found that semi-natural grassland has greatly declined in area since 1945, with losses of around 90\% in the UK's lowlands. By 2011 only 2\% of this grassland area was classed as \textit{high diversity} with two studies showing that a loss of 97\% of enclosed semi-natural grassland was seen between 1930 and 1984 \citep{LotV1, UKNEA2011}, mainly due to agriculture improvement \citep{warzecha2018attractiveness}.
The Countryside Survey 2007 \citep{CEH2007} showed that the decline had slowed between 1998 and 2007 allowing the focus to be shifted toward increasing the amount of grassland and the diversity within. 50\% of all grassland plant species can be identified on road verges, including 91 species which are either threatened or near threatened \citep{Plantlife2017}. 

Surveying and classifying verges is not only a time consuming activity that requires experts and volunteers alike, but often a legal requirement. Local Wildlife Sites (LWSs) are designated areas identified as having local conservation value, and serve to protect local diversity while complementing a more general environmental stewardship. Local authorities are required to identify LWSs based on indicators of local biodiversity, and as a way of assessing the effectiveness of their planning systems \citep{neil_local_2013}. Diversity within grassland also requires management as part of the response to the climate emergency. Species rich grassland stores more carbon than species poor grassland with important interactions between, and within, species  \citep{govwales}. Management of verges is critical for these interactions, but also for authorities and businesses who will be required to demonstrate \emph{biodiversity net gain} due to UK government policies aimed at mitigating climate change.

An important factor within verge management is the role of pollinators, specifically within roadside verges, but also within agriculture in general. Pollinators and wildflowers can be functionally linked, with a decline in one seeing a decline in the other \citep{biesmeijer2004parallel} which contributes to the importance placed upon diversity. Diversity and its management was addressed with a review of 140 relevant studies, providing recommendations for verges and entire road networks. Pollination and diversity must go together when considering grassland management due to a lack of holistic and large-scale understanding of the net effects that roadside verges have upon pollinators \citep{PHILLIPS2020108687}. Species rich verges, and more generally \emph{green space}, also contribute on a psychological, and general well-being level to the public \citep{stubbings_hierarchical_2019, osullivan_optimising_2017}.


New and efficient ways of managing verges, taking advantage of technological improvements, are necessary if conservation efforts are to succeed over such a large area of grassland. The use of imagery within ecology studies is an often overlooked source of data  \citep{depauw_use_nodate}, due to the lack of quantitative methods to estimate the presence and extent of different species of flora and fauna. Computer vision, and more recently artificial intelligence (AI), can be used to extract meaning from unstructured images in a way that is not possible for a human. AI can efficiently automate the identification of scene content within imagery (e.g., wildflower species) and provide a new way to help naturalists manage verges, producing what has been termed an ``AI Naturalist'' \citep{AINaturalists}. Such an approach brings together the latest state-of-the-art machine learning technologies, such as artificial neural networks (ANN), and large publicly available image datasets, such as Google Street View (GSV) \citep{GSV}, to provide a monitoring and surveying technique that ecologists \emph{could} take advantage of.


In this work, we propose an AI-based approach to automated surveying of road verge habitat quality using convolutional neural networks (CNNs) and publicly-available street-view imagery of roadside verges. Utilising a supervised learning approach, a state-of-the-art CNN was trained on thousands of images with corresponding ground truth labels obtained from one of the UK's largest road verge surveys, ``Life on the Verge''. Our approach, entitled \textit{DeepVerge} is, to our knowledge, the first application of AI to ecological conservation based on street-view imagery.


\section{Related Work}
\label{sec:relatedwork}
\subsection{Convolutional neural networks}
Convolutional Neural Networks (CNNs) \citep{oshea_introduction_2015} are continually pushing boundaries in terms of image classification. Examples of recent state of-the-art (SOTA) CNNs include ResNet \citep{he2016deep}, EfficientNet \citep{EfficientNet} and ResNeXt \citep{xie_aggregated_2017}, to name just a few. A CNN has two main parts: a convolutional backbone for representation learning, and a fully connected neural network (NN) that serves as a classifier. CNNs can be greatly enhanced through pre-training on a larger dataset (possibly from a different domain altogether) and harnessing the learned representations in a process called transfer learning \citep{perez_effectiveness_2017}. Massive labelled image datasets \citep{NIPS2014_b86e8d03} are routinely used to train CNNs, allowing them to learn information that is commonly found within many image domains. Applications to domains with limited data thereby benefit from reduced training times and/or increased performance in not having to learn these common traits from scratch. There is mounting evidence that these algorithms can just be taken \emph{off the shelf} \citep{DBLP:journals/corr/RazavianASC14} and then fine-tuned to specific datasets.

\subsection{Street-view imagery}
A review of street view imagery for urban analytics found over 600 papers using this imagery and that GSV is an entrenched component of urban analytics and GIScience \citep{BILJECKI2021104217}. One of the earliest examples of GSV use, for remote surveying,  \citep{FriedlandPhdthesis} used street level imagery to compare building damage before and after a hurricanes Katrina and Ike. Possibly the first remote ecological study to make use of GSV  \citep{ROUSSELET2013} looked at the species distribution of the Pine Processionary Moth. GSV images were used to infer the presence of the moth and derive an occurrence metric. The results showed that their derived metric correlated highly with the field study achieving an accuracy of 96\%.
\par
Evaluation of GSV roadside verges were used for tracking invasive alien plants along roads \citep{kotowska_evaluating_2021} and its use was found to have a great potential to support ecological research in documenting species distribution. While both \citet{ROUSSELET2013} and \citet{kotowska_evaluating_2021} used humans in the loop to visually detect occurrence within the GSV imagery, \citet{RINGLAND201936} used a CNN to detect crop types within the imagery and even automated the collection of GSV images. Humans were still used in their pipeline however, within the selection of suitable images, and they achieved an overall accuracy of 83\% classifying crop types. \citet{deus_google_2016} compared the more common method of using car surveys (CS) with what they termed as Remote Sensing using GSV as a possible cost effective alternative. They, like others, used GSV to determine occurrence of a single plant species (Eucalyptus globulus Labill) using human observers, and while finding lower occurrence using GSV they were able to improve the results using other variables. They went on to conclude that GSV use was a cost-effective alternative to CS.
\par
While the combination of GSV and CNNs has been used in contexts such as herb identification \citep{zhao_cnn_2020} and street sign detection \citep{campbell_detecting_2019}, the combination typically focuses on specific object detection or classification \emph{within} the scene, rather than classifying the scene itself. Scenery classification has been made within the urban environment \citep{nguyen_using_2020}, where the presence of crosswalks, building types, utility wires, single lane road and a percentage of greenery were used to classify COVID-19 risk of neighbourhoods. The \citet{nguyen_using_2020} study still used human in the loop for annotation of ground truth and also directed the CNN toward object indicators. Our study, by comparison, automates annotation without directing the CNN toward any specific object.     

\section{Data Preparation and Curation}
\label{sec:data}
\subsection{Ground truth}
\begin{figure}[t!]
    \centering
    \captionsetup{width=0.6\textwidth}
    \setlength{\fboxrule}{1pt}%
        \fbox{\includegraphics[width=0.6\textwidth]{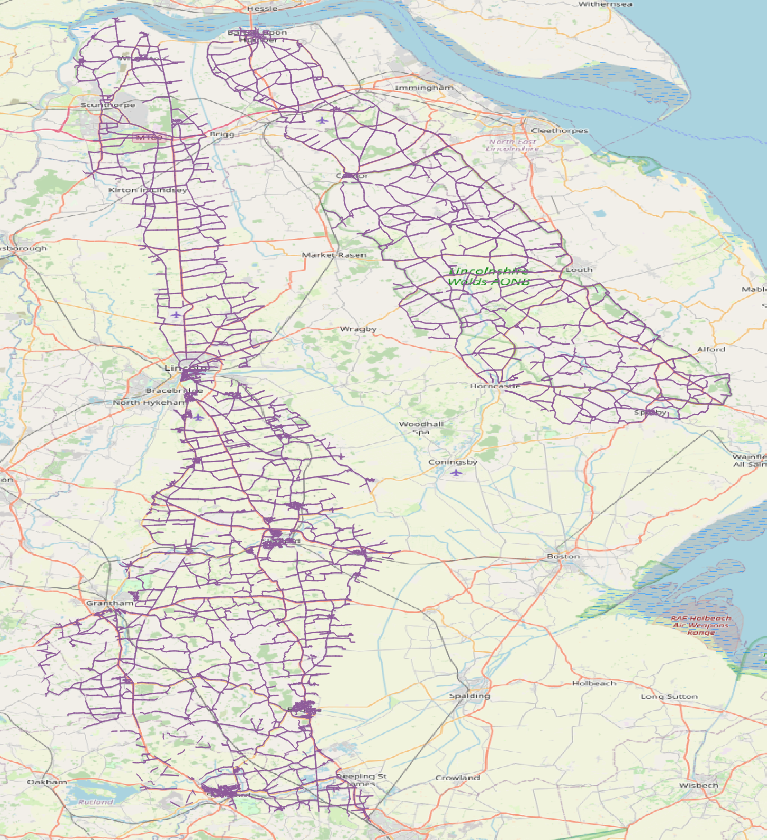}}
        \caption[Overview of LWT LotV Survey Area]{Overview of LWT LotV Survey Area. Image produced by the Open Source application QGIS \\ (\url{https://www.qgis.org/en/site/})}
        \label{fig:KML-Overview}
\end{figure}

Lincolnshire Wildlife Trust (LWT) has conducted thorough surveys of Lincolnshire's roadside verges under a project entitled ``Life on the Verge (LotV)''. The project included three citizen science sub-projects that correspond to 3,500 road sections/specific localities of Lincolnshire Wolds (LotVW), Lincolnshire Northern Edge, and Limestone Grassland (Fig.~\ref{fig:KML-Overview}). The results of these surveys provide ground truth (GT) for this study.
\par
Each section has the roadside verges within scored by a value corresponding to the number of positive indicator species identified \citep{neil_local_2013}. These scores were later quantized into a LotVW ordinal scoring system with five levels, as detailed in Table. \ref{tbl:categories}, describing to the verge's biodiversity status and conservation potential. The 5\textsuperscript{th} category is special, in that it includes the most important areas having already been designated as Roadside Nature Reserves (RNRs) \citep{BIOTT2013} (LWT Internal Document).

\addvbuffer[12pt 8pt]

\begin{table}[H]
\captionsetup{width=12cm}
\centering
\scriptsize
    {%
    \noindent\begin{tabular}{c|c|c|c}
        \textbf{} & \textbf{} & \textbf{Positive Indicator} & \textbf{Number of}\\
        \textbf{Class} & \textbf{Description} & \textbf{Species (Score)} & \textbf{Examples} \\
        \hline\hline
        5 & Roadside Nature Reserves & 20+ & \\
        4 & High Local Wildlife Site & 12+ & \\
        3 & Local Wildlife Site & 8 to 11 & \\
        2 & Has Conservation Potential & 4 to 7 & \\
        1 & No Conservation Potential & 0 to 3 & \\
        \hline
    \end{tabular}
    }
\caption{Ground Truth Survey Scores to Category Mapping}
\label{tbl:categories}
\end{table}

Surveying started in 2009 with the first LotVW sub-project \citep{LotV2013} involving 1,100 hours from over 100 people to cover 1,115 km of roads. Between 2009 and 2016, surveying was expanded to 3,900 km of roadside verges and found 159 new LWS of wildflower-rich habitat \citep{LotV1}. The mean road length, for each section, was approximately 1 km with the number of GPS-derived GT locations in each varying from 3 to 30+. Overall the survey results provided approximately 50,000 GPS locations that require mapping or \emph{snapping} to their associated GSV panorama locations to create a usable GT.

Any road has features such as junctions, bends and straights, which themselves have features such as the start, middle and end. A road section therefore can be \emph{described} with a series of GPS locations placed at each of these features. These road feature locations, along with 8-way compass directions identifying the number of wildflower species of each verges section are provided within the LWT survey results.  

\subsection{Raw data}
\label{sec:curation}
\begin{figure}[t!]
    \centering
    \captionsetup{width=1\textwidth}
        \includegraphics[width=\textwidth]{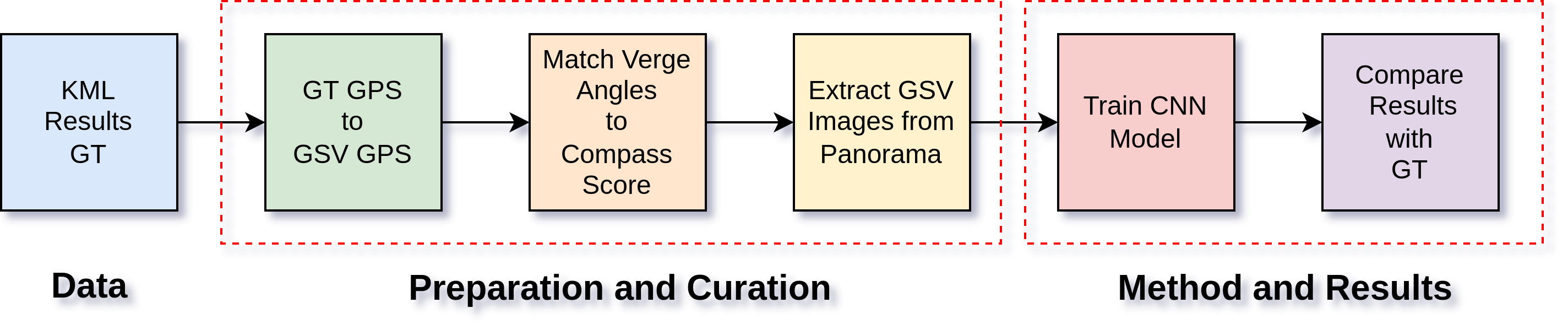}
        \caption[Data Pipeline]{Data Pipeline for Training a CNN using LWT Survey Results }
        \label{fig:pipeline}
\end{figure}

With the aim of developing a remote means of surveying roadside verges, the ability to identify new LWSs was identified as a core objective. In order to train a model with this capability however, it was first necessary to transform the raw data into a useable dataset. The survey data was provided by LWT in Keyhole Mark-up Language (KML) \citep{OGC1}, which required processing in order to produce a usable GT annotated image dataset. Additionally, imagery required extraction via an Application Programming Interface (API) from within 360{\degree} GSV panoramic images taken by Google over the last decade or more. The pipeline (Fig. \ref{fig:pipeline}) was almost entirely automated and began by parsing the KML file and translating the GT GPS coordinates that describe the road approximately, into GSV panorama GPS locations, which describe the road accurately. The panorama locations are an accurate representation of the road since they were generated along the path of the Google car.
\par
GSV panoramas were investigated to find their spacing and coverage, which varied from 10 to 20 metres \citep{mazerolle2011google}. The angle between sequential panorama locations gave the bearing of the road, which in turn revealed the perpendicular angle to the verge, allowing score label(s) to be assigned to each location. Providing that the location's verge direction had a corresponding score within the survey data, images were extracted from the panorama. The collection of these images became the annotated dataset, which was later used to train the CNN.
This process also highlighted several issues. In particular, survey GPS locations often failed to line up with GSV panorama GPS locations, and could easily snap to a panorama within a different road, especially around junctions, with a naive approach. In addition to this, if there were not enough GT locations describing the road, or their accuracy was poor, it would lead to incorrectly calculated angles and cause extracted images to be mislabeled (Fig. \ref{fig:KML-RoadFeatures1}).

\subsection{Road features and panorama locations}
\label{subsec:roadfeatures}
A small change in the number of GPS locations describing a road section can have a large effect on how these locations are labelled using the GT. As an example, adding a location C in Fig. \ref{fig:KML-RoadFeatures1} results in a more faithful orientation being assigned to the road section between locations A and B.

\begin{figure}[t!]
    \centering
    \captionsetup{width=1\textwidth}
        \includegraphics[width=\textwidth]{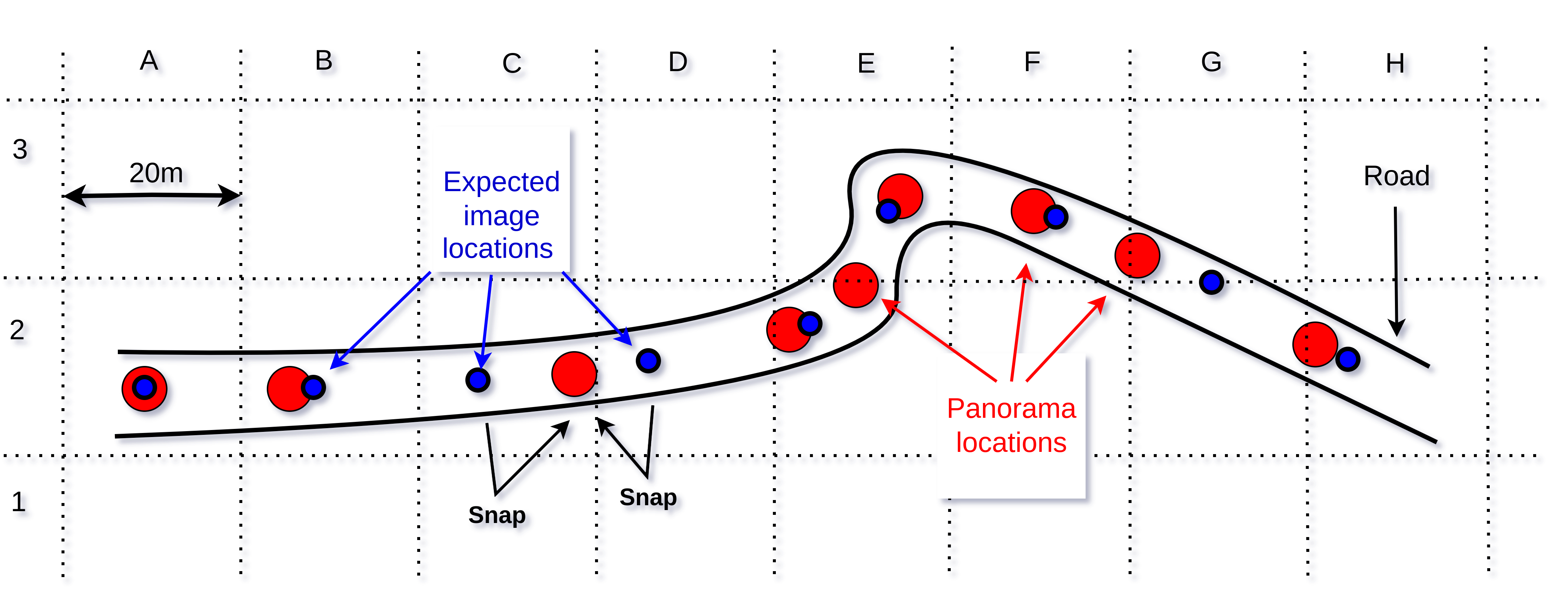}
        \caption[Extracting Images Via A Grid]{Extracting images via a grid system or extracting every x metres - can cause problems as the expected image will \emph{snap} to the nearest panorama centre.}
        \label{fig:RoadGrid}
\end{figure}

When annotating CNN training images, accidental mislabelling introduces noise within the GT dataset. Such noise can originate from duplication (Fig. \ref{fig:RoadGrid}), incorrect verge angles (Fig. \ref{fig:KML-RoadFeatures1}), snapping to the wrong panorama, or from the natural species variability of roadside verges. Interpolation of panorama locations between two points, (such as Point A and C, Fig. \ref{fig:KML-RoadFeatures1}) may reveal a further set of GSV panoramas in which to extract verges images from.

Not all of the images extracted from panoramas will be good examples of the score associated with the compass octant direction. The natural variability along any single verge section may see patches of high species rich grassland and, very likely, also areas with poor conservation potential, yet all images from this section will be labelled with the same score. 

\begin{figure}[t!]
    \centering
    \captionsetup{width=0.8\textwidth}
        \includegraphics[width=0.8\textwidth]{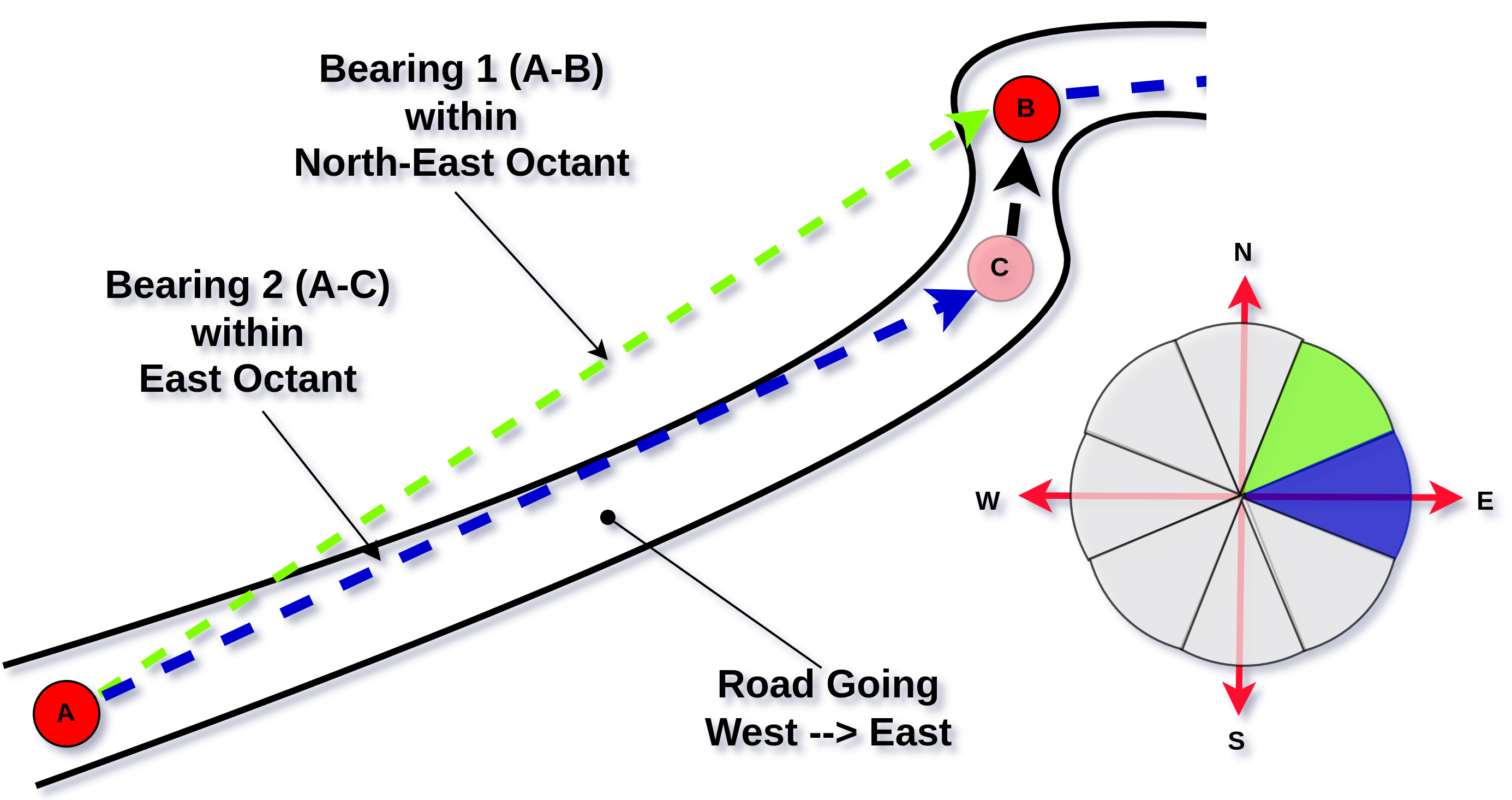}
        \caption[Investigating Road Features]{Investigating Road Features - more points describe the road more accurately.}
        \label{fig:KML-RoadFeatures1}
\end{figure}

Various researchers \citep{stubbings_hierarchical_2019, kim_decoding_2021, nguyen_using_2020} discuss using a grid, or intervals, to choose where to take images from. However, with the panorama locations having variable spacing this can lead to duplicate images being captured due to the nature of the GSV \emph{snap to grid} translation of GPS points to panorama points.

While requesting images from the expected GPS locations (Fig. \ref{fig:RoadGrid}) the requested image from D2 will return the same panorama as for C2. The Fig. \ref{fig:RoadGrid} example will lead to an 11\% image duplication which can cause the CNN to favour features within these duplicated images. 

\subsection{GT GPS Translation to GSV}
\label{subsec:translation}
\begin{figure}[t!]
    \centering
    \includegraphics[width=\textwidth]{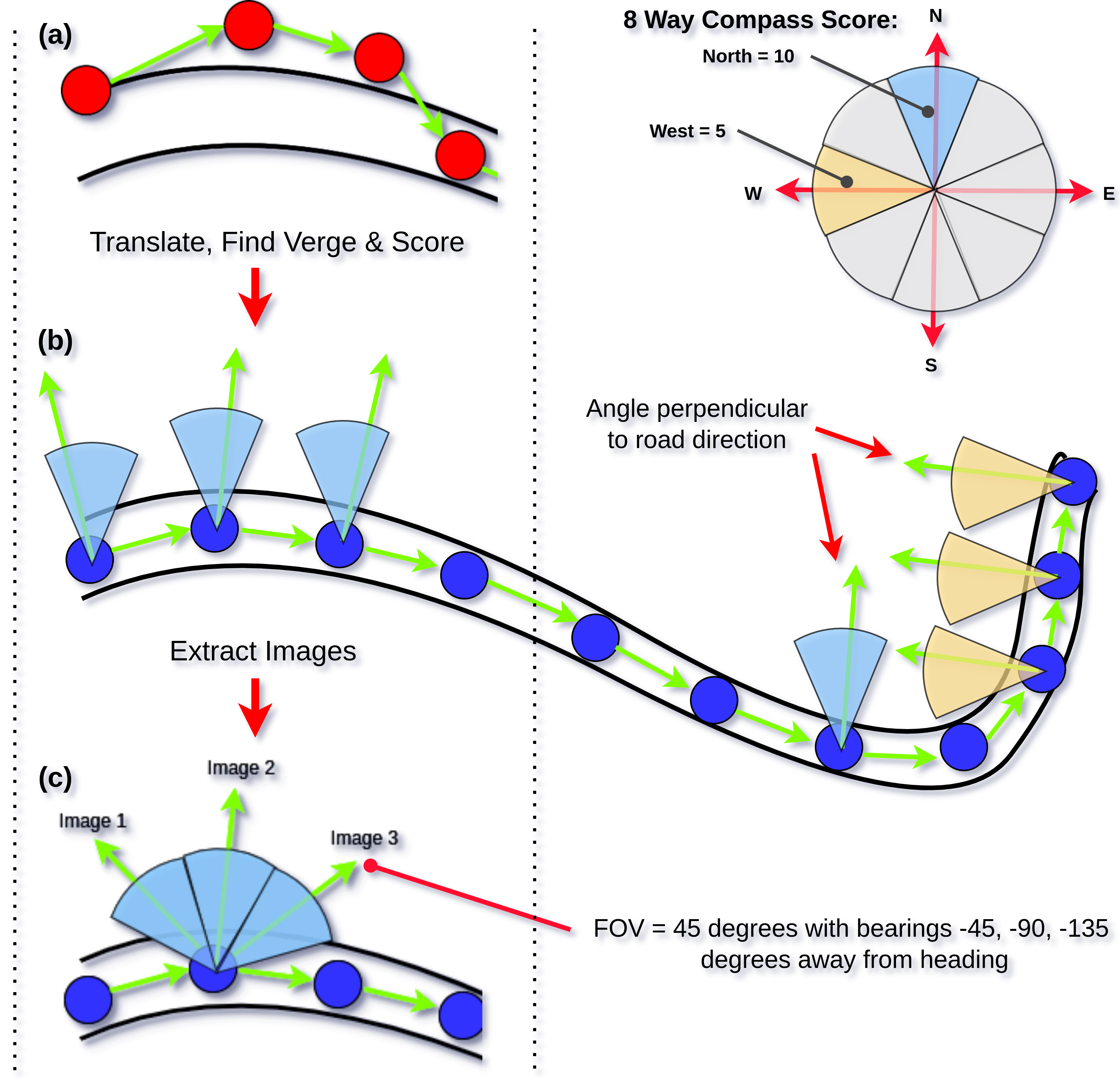}
    \caption[LotV GPS Coords Transformation]{LotV GPS Coords Transformation \\(a) Original LWT GPS (b) Translated to GSV panoramas and compass scores matched with perpendicular verge (c) Images extracted and labelled}
    \label{fig:GSV-Transform} 
\end{figure}

Having investigated GPS locations that describe road features and their correlation to GSV panorama locations, a translation from one to the other was performed (Fig. \ref{fig:GSV-Transform}a). Translation allows matching GT 8-way compass scores to the perpendicular verge at these translated panorama locations (Fig. \ref{fig:GSV-Transform}b). Any panorama location that correlates to a GT score can have images extracted (Fig. \ref{fig:GSV-Transform}c) to build the GT dataset. 

\subsection{Image Extraction and Dataset Building}
\label{subsec:extraction}
A decision needed to be made in regards to how much and exactly what part of the panorama will be used. Fig. \ref{fig:GSV-Transform}b shows two matching scoring directions of North and West with 3 other locations failing to be matched as their verge angle falls outside of a scoring octant. Extracting an image with a Field Of View (FOV) of 180{\degree} perpendicular to the road would capture the entire verge, however this wide angle would also capture a lot of sky and road. Instead, three images of 45{\degree} are taken. The FOV equates to \emph{zooming}. Zoom in and the image can become pixelated, zoom out and details are lost.

\begin{figure}[t!]
    \centering
    \includegraphics[width=\textwidth]{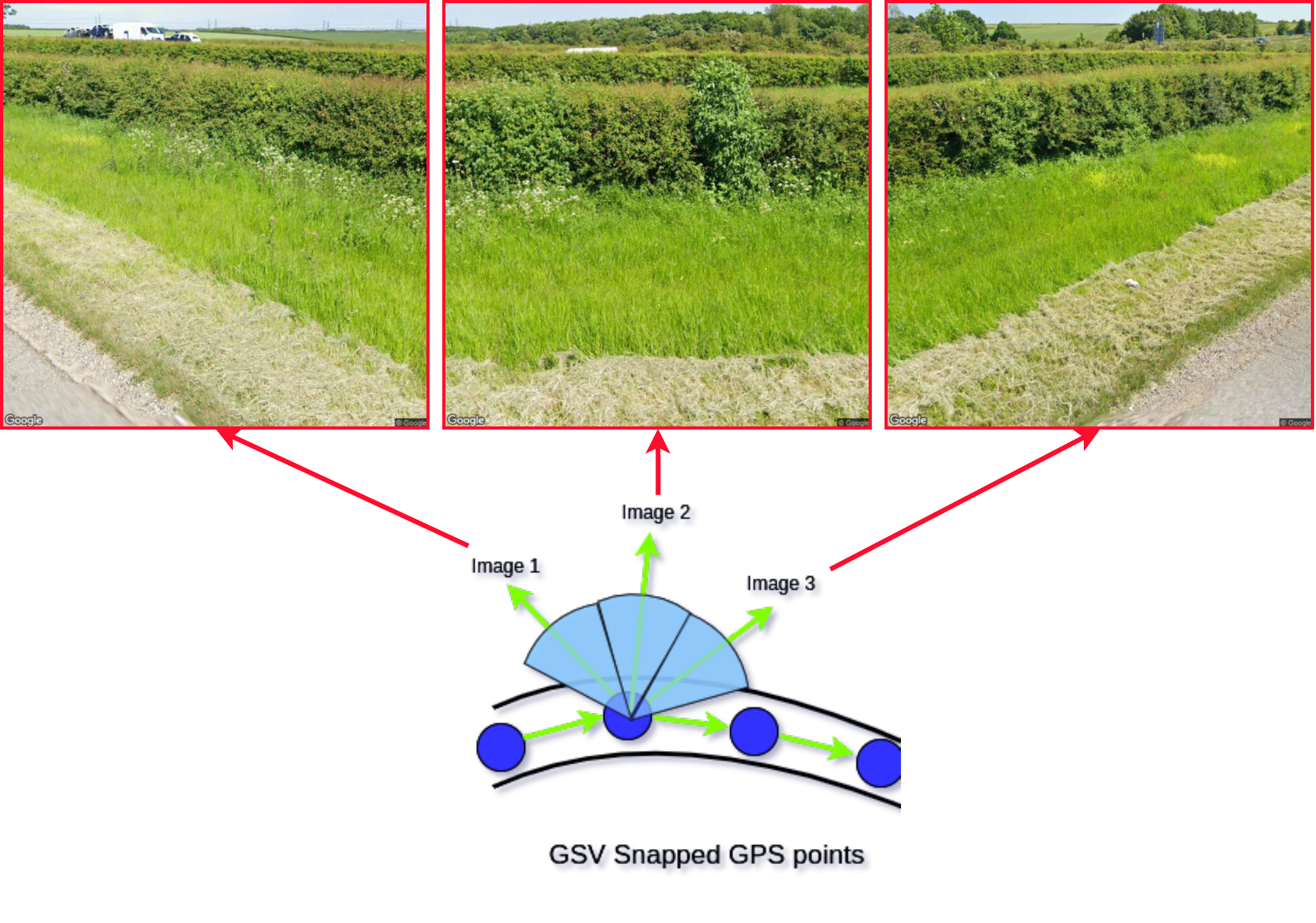}
    \caption[Obtaining Images]{Obtaining Images}
    \label{fig:GSV-Images2}
\end{figure}

A FOV of 45{\degree} was chosen as a good compromise that also correlates with an 8-way compass (Fig. \ref{fig:GSV-Images2}). A pitch of 20{\degree} was applied (moving the camera up within the panorama) to reduce the amount of road in the images.

When building the dataset additional criteria was used to allow finer control over which images were chosen in order to reduce noise. Firstly, images were extracted from the 2009 summer months (June,July and August) from within the \emph{Lincolnshire Wolds} area, to match the LovW survey data. This produced 3396 images, with scores ranging from 0 to 19 but a 3rd of these fell into the lowest category. To increase the number of images the criteria was modified to also include the \emph{Northern Lincolnshire Edge} area as well as the year 2021. This produced 6882 images with more images falling within other categories. GSV added images for both these areas in 2009 and 2021, with the later year also providing better quality images. All images were downloaded using the GSV API and have a size of 640 by 640 pixels, many of which have the Google logo embedded within.

A manual annotation of these images was carried out to mark and remove \emph{bad images}, which we classed as images that included cars, houses, verges that had been completely cut or where the verge was not in view. This removed 889 images. Due to road sections having start and end points, some sections shared the same panorama locations and produced duplicate images. 44 Duplicate images were removed leaving 5949 images within the dataset. There will still be noise present within the dataset, represented by poor quality imagery and images that mainly contain hedges or fields.

Images were not taken from the $3^{rd}$ area of \emph{Limestone Grassland} so that there are images from an area that the model has never seen before, which will allow future model performance to be tested within a different locality.

\section{Methodology}
\label{sec:method}

\subsection{Baseline model}
\label{subsec:featureextraction}
\begin{figure}[!tb]
    \centering
    \includegraphics[width=\textwidth]{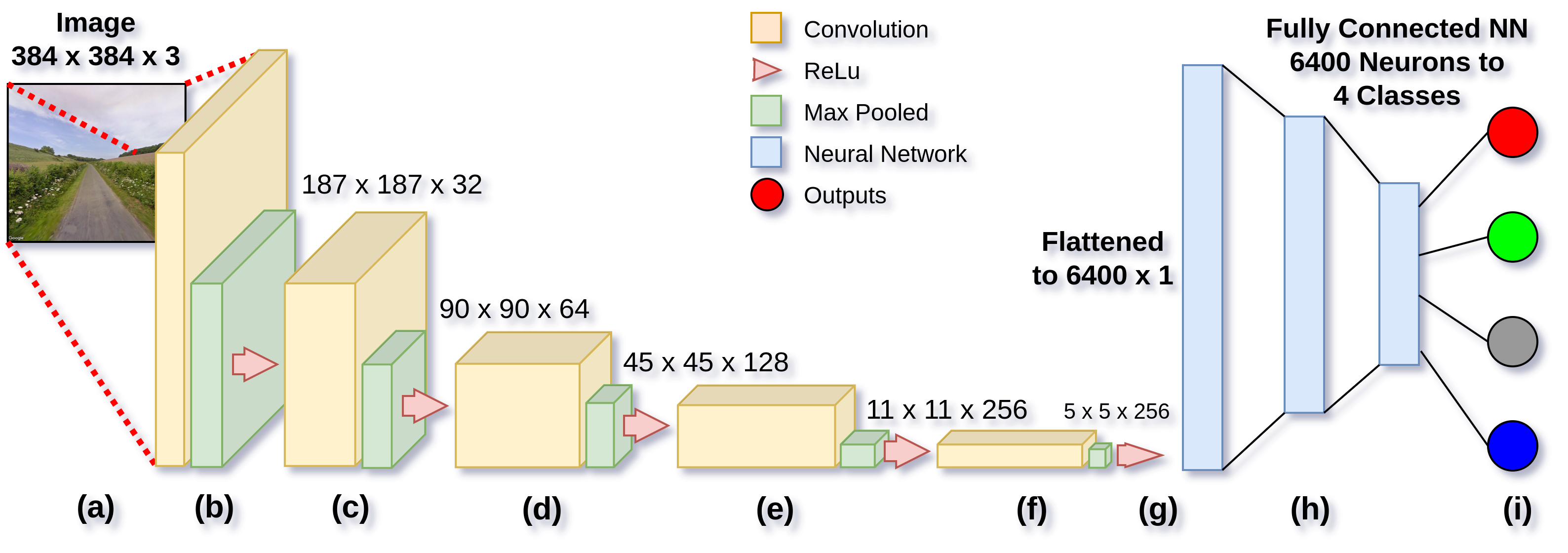}
    \caption[OurNet]{Baseline CNN model \\(a) RGB images augmented and transformed (Sect. \ref{subsec:augmentation}) feed into (b) $1^{st}$ Convolution block using a $11^{2}$ kernel (c) $2^{nd}$ Convolution block using $7^{2}$ kernel (d) $3^{rd}$ block is repeated twice using kernels of $5^{2}$ (e) $5^{th}$ block has kernel of $3^{2}$ (f) The final convolution block also uses a $3^{2}$ kernel and outputs 256 feature maps (g) Flattens the array to 6400 x 1 which feeds into (h) the FCNN with a 1 to 1 mapping between feature maps and neurons with 2 hidden layers (i) 4 FCNN outputs of classifier}
    \label{fig:OurNet}
\end{figure}

The key role for the use of a CNN is to find and extract objects within images, $1^{st}$ stage, that belong to individual classes so that classification, $2^{nd} stage,$ can be made. A baseline CNN model (Fig. \ref{fig:OurNet}) was built that could be used for experimentation, and to compare the effectiveness of employing a SOTA CNN model to improve performance. 


The baseline model comprises of 6 convolution blocks (Fig. \ref{fig:OurNet}b-f, with d repeated), with each block followed by batch normalised \citep{ioffe_batch_2015}, rectified with ReLu non-linearly \citep{agarap_deep_2019} and max pooling \citep{sun_ampnet_2021}. Each block progressively increases the number of feature maps while reducing their dimensions. The Convolutions take the original 3 channel RGB image of 384 x 384 pixels down to 256 channels of 5 x 5 pixels, known as down sampling (\cite{shorten_survey_2019}). The resulting feature map array has a depth of 256 which is then flattened before being fed into the FCNN. The flattening allows connection of the 256 x $5^{2}$ feature maps to 6400 artificial neuron inputs of the FCNN. Three layers or neurons are batch normalised followed by ReLu non-linearity before outputting the 4 classes (Fig. \ref{fig:OurNet}g-i).

The difference between the known class, the GT, and the learned probable class produces an error metric. A proportion of the error is back propagated \citep{li_brief_2012} through the model to update the filter parameters so as to reduce the error on the next training run (epoch). After multiple epochs this error is minimised and thereby classifying each image.

The process learns the best filter parameters that describe features within the image that are spatially independent from both the images themselves and other features found. 

Any parameter, a variable, that is set before the model is run to control its behaviour is called a hyper-parameter, as opposed to parameters that get changed as a result of the running. Hyper-parameters were tuned during training (Sect. \ref{subsec:training}).

\subsection{Augmentation and Image Transformations}
\label{subsec:augmentation}

\begin{figure}[!t]
\captionsetup{width=10cm}
    \centering
    \includegraphics[width=0.75\textwidth]{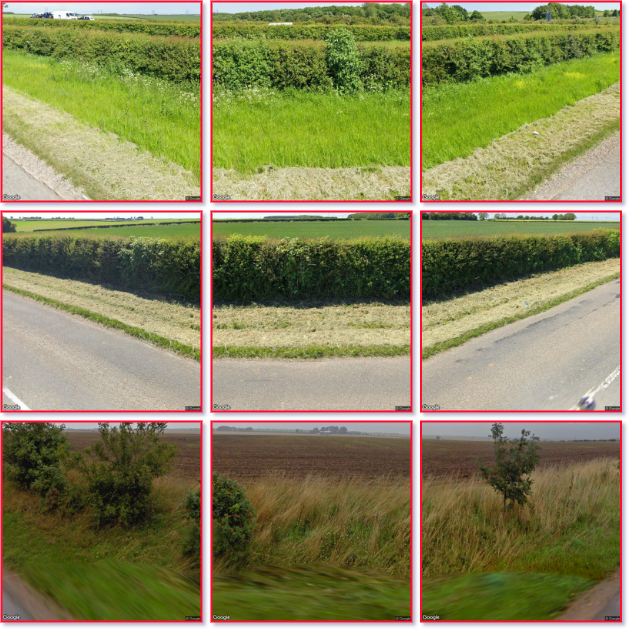}
    \caption[Real variation of image samples]{Real-world variation of image samples, dependent on the position of the camera with respect to the verge. These variations were used to inform the augmentation policy used.}
    \label{fig:Augment}
\end{figure}

Augmentation is used to both artificially increase the number of images and to make random image transformations of each image, since the model needs to learn generalised features and not precise features \citep{shorten_survey_2019, li_simple_2021}. Since the Google Car (in most cases) drives on one side of the road and not in the middle, extracted images take on perspective, rotation, brightness and scale variations. Fig. \ref{fig:Augment} shows images from 3 panoramas highlighting these variations and can be used to inform an augmentation policy. The variations show perspective, brightness and contrast and scaling differences, leading to an augmentation policy that can include images being randomly horizontally flipped, normalised, scaled and with perspective distortion. Additionally, it can be assumed that the car will go over bumps, effectively causing possible camera rotation.

Any given image will be fed into a CNN model multiple times (once per epoch / training run) but transformed randomly each time, so the model sees multiple variations of the same image. Features within the images that remain invariant to the random transformations will be learnt. With each image having a classification label (a score group) situations arise whereby some classes (groups) have many more image examples in than others (class imbalance). The imbalance between classes can be evened out by over-sampling minority classes to match the majority class. Over-sampling involves duplicating images but applying one or more random transformations (Table \ref{tbl:tranformations}) to each image to form a different image. 

\addvbuffer[12pt 8pt]

\begin{table}[t!]
\captionsetup{width=12cm}
\centering
\scriptsize
    {%
    \noindent\begin{tabular}{c|c|c|c}
        \textbf{Transformation} &
        \textbf{Description} & \textbf{p} & \textbf{Parameters} \\
        \hline\hline
         Random Flip & Image is flipped horizontally & 0.50 & \\
         Random Erasing & Occlude a section of the image & 0.70 &\\
         Random Noise & Adds salt and pepper noise & 0.30 & 10\% coverage \\
         Image Crop & Randomly select an area and resize & 0.50 & random sizes \\
         Random Perspective & Distort image perspective & 0.50 & distortion up to 50\% \\
         Random Rotation & Rotate image anti- or clockwise & 0.50 & -5 to +5 degrees \\
         Normalisation & Zero-mean, unit-variance  & 1.00 & \\
         Downscale Image & Set all images to 384 x 384 pixels  & 1.00 & \\
         Top \& Bottom Crop & Remove part of sky and road  & 1.00 & top 174px, bottom 40px \\
        \hline
    \end{tabular}
    }
\caption{Image transformations that were tried as part of an augmentation policy. Either individually or as a collection of transformations.}
\label{tbl:tranformations}
\end{table}

\subsection{Training and Experiments}
\label{subsec:training}
Training the model involved splitting the dataset up into training, validation and test (held-out) sub-sets (70\%, 10\% and 20\%). The validation set was augmented and balanced and was used for hyper-parameter tuning during training runs and the held-out set used for measuring the model's performance on images not seen by the model, to simulate a more \emph{real world test}.

Each training run consisted of 50 or 100 epochs with each epoch presenting the training set to the model using a different image shuffle, along with a different random augmentation applied. The shuffling and random augmentations require controlling via the use of Python's, Pytorch's and Numpy's Random Number Generator (RNG) \citep{REPRODUCE} seed in order for experiments to be reproducible. A seed of 0 was used throughout. 

A software tool was developed so that the entire pipeline from parsing the LWT KML file to the choice of CNN hyper-parameters, approximately 75 variables, could be configured and controlled. With such a large number of variables a full grid search seeking the absolute best parameters could not be employed, however a manual grid search was employed concentrating upon the main hyper-parameters of the CNN and covered approximately 500 training runs.  

Experiments were conducted using Ubuntu 21 OS with 32GB main memory, Intel i9 12\textsuperscript{th} Gen. processor and a Nvidia RTX 3080 Ti GPU. 

Once the most important tuning parameters were found, various experiments were carried out by iteratively adding different augmentations and other components (Table. \ref{tbl:iterations}), changing the model each time, and looking at the effect each had upon the model's performance, as measured by accuracy in predicting the class of each image.

\begin{table}[t!]
\captionsetup{width=10cm}
\centering
\scriptsize
    {%
    \noindent\begin{tabular}{c|c|c|c|c}
        \textbf{} & \textbf{} & \textbf{Epoch} & \textbf{Validation} & \textbf{Test} \\
        \textbf{Iter.} & \textbf{Addition To Model} & \textbf{Time} & \textbf{Accuracy} & \textbf{Accuracy} \\
        \hline\hline
        0 & Image $640 \times 640$ resize to $224 \times 224$ & 0m 28s & 68.90\% & 67.44\% \\
        1 & Transform random h-flips & 0m 29s & 73.11\% & 71.29\% \\
        2 & Bad Image Purge & 0m 26s & 73.79\% & 70.98\% \\
        3 & Over-sampling class balancing & 0m 55s & 72.62\% & 69.81\% \\
        4 & Random perspective and rotation & 2m 25s & 78.96\% & 74.98\% \\
        5 & Image Resize to 384 x 384 & 2m 29s & 79.80\% & 76.73\% \\
        6 & Optimisations AMP, PIN \& CLMF & 2m 4s & 79.63\% & 77.81\% \\
        7 & Fine-tuning at lr=1e\textsuperscript{-05} & 2m 7s & 84.14\% & 80.57\% \\
        8 & Batch size from 32 to 128 \& -CLMF & 1m 58s & 83.64\% & 80.82\% \\
        9  & Random erasing & 2m 03s &  82.52\% & 82.27\% \\
        \hline
    \end{tabular}
    }
\caption{Iterative progression of baseline model (Seed 0)}
\label{tbl:iterations}
\end{table}

Each iteration, 0 to 6, was trained over 50 epochs with iterations 7 to 9 fine tuned with a further 50 epochs from iteration 6. As components are added some early iterations produced worse results, such as adding the component \emph{bad image purge}, which turned out to be caused by duplicate images being left within the dataset. This was rectified in the last iteration. The benefit of over-sampling to balance the classes was not seen until further augmentation transforms were applied in later iterations. 

Gradually a model and pipeline was arrived at that gave a result of 82.27\%, tested against the held-out image set. The result suggested that the augmentation policy and hyper-parameters were ready to be tested using SOTA models to achieve a performance increase.

The baseline model was substituted for SOTA CNN models of increasing depth. The training results (Table \ref{tbl:models}), showed that pre-trained SOTA models gave upto a 6.72\% test accuracy improvement, with no other alterations to the pipeline. The models have been pre-trained using ImageNet challenge \citep{ILSVRC15} and therefore give the model a lot of prior knowledge as a starting point, with our pipeline fine tuning the model for our specific need. The image batch size was set to maximise available memory, nearest power of 2, with the final layer of the FCNN, which has 1000 classification outputs, reduced to 4. 

\begin{table}[tb!]
\captionsetup{width=12cm}
\centering
\scriptsize
    {%
    \noindent\begin{tabular}{c|c|c|c|c|c}
        \textbf{} & \textbf{Epochs} \& & \textbf{Batch} & \textbf{Epoch} & \textbf{Validation} & \textbf{Test} \\
        \textbf{Model} & \textbf{Learning Rate} & \textbf{Size} & \textbf{Time} & \textbf{Accuracy} & \textbf{Accuracy} \\
        \hline\hline
        ResNet-18 & $50 \times 1e^{-4}$ & & &  \\
        no pre-training & FT $\longrightarrow$ $50 \times 1e^{-5}$ & 128 & 1m 51s & 84.87\% & 80.50\% \\
        \hline
        ResNet-18 & $50 \times 1e^{-4} - 50 \times 1e^{-5}$ & 128 & 1m 53s & 87.56\% & 86.13\% \\

        ResNet-34 & $50 \times 1e^{-4} - 50 \times 1e^{-5}$ & 128 & 1m 49s & 88.91\% & 86.97\% \\

        ResNet-50 & $50 \times 1e^{-4} - 50 \times 1e^{-5}$ & 64 & 1m 49s & 88.24\% & 86.55\% \\

        ResNet-101 & $50 \times 1e^{-4} - 50 \times 1e^{-5}$ & 48 & 1m 49s & 89.58\% & 87.73\% \\

        ResNet-152 & $50 \times 1e^{-4} - 50 \times 1e^{-5}$ & 32 & 1m 51s & 89.24\% & 88.91\% \\
        \hline
        ResNeXt-50 & $50 \times 1e^{-4} - 50 \times 1e^{-5}$ & 32 & 1m 51s & 90.76\% & 88.99\% \\

        ResNeXt-101 & $50 \times 1e^{-4} - 50 \times 1e^{-5}$ & 16 & 2m 36s & 89.75\% & 87.73\% \\
        \hline

        EfficientNet-B0 & $50 \times 1e^{-4} - 50 \times 1e^{-5}$ & 32 & 2m 22s & 87.90\% & 86.81\% \\        
        EfficientNet-B4 & $50 \times 1e^{-4} - 50 \times 1e^{-5}$ & 24 & 5m 37s & 89.75\% & 86.81\% \\ 
        \hline
    \end{tabular}
    }
\caption{Progression of Pre-Trained SOTA Models (Seed 0). Trained over 50 epochs then fine tuned (FT) for a further 50 epochs}
\label{tbl:models}
\end{table}%

The deeper the model the harder it will be to train \citep{he2016deep}, although this has not been experienced. The first ResNet-18 model had no previous knowledge before training but still had a similar number of learnable parameters, compared to the baseline, but did not perform as well, possibly being caused by the model being too complex to train with limited epochs.

\begin{figure}[tb!]
\captionsetup{width=6cm}
    \centering
    \begin{minipage}[t]{0.5\textwidth}
        \centering
        {%
        \setlength{\fboxrule}{1pt}%
        \includegraphics[width=0.95\textwidth]{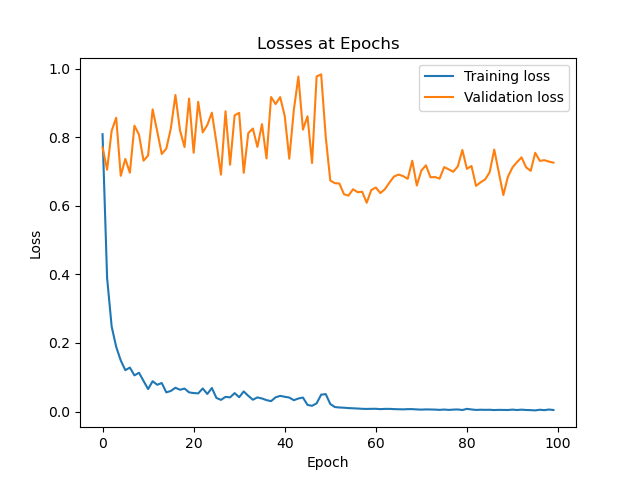}}%
        \caption[Training and validation loss]{Training and validation loss of ResNeXt-50 model for seed 0} 
        \label{fig:accuracy}
    \end{minipage}\hfill
    \begin{minipage}[t]{0.5\textwidth}
        \centering
        {%
        \setlength{\fboxrule}{1pt}%
        \includegraphics[width=0.95\textwidth]{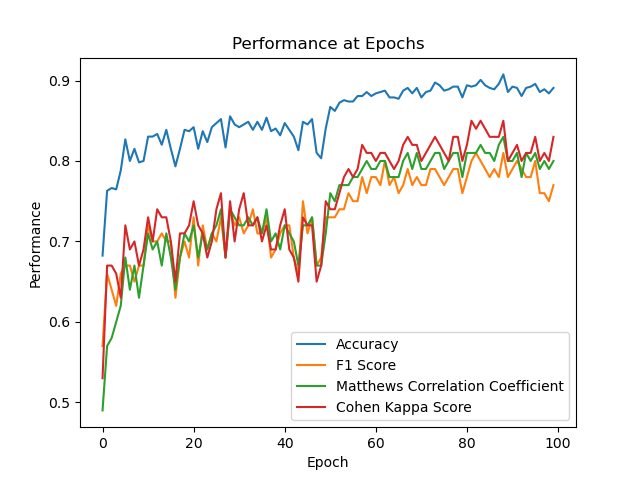}}
        \caption[Various Performance Metrics]{Validation performance metrics of ResNeXt-50 model for seed 0. F1-Score uses macro averaging and Cohen Kappa uses quadratic weighting}
        \label{fig:performance}
    \end{minipage}
\end{figure}

Training performance plots, (Figs. \ref{fig:accuracy} and \ref{fig:performance}), clearly show the end of training using a learning rate of 1e\textsuperscript{-04} for the first 50 epochs. The learned knowledge is then passed back into the model for the second stage of training at the smaller learning rate of $1e^{-5}$.

The performance metrics are all calculated using multi-class versions of standard binary class metrics. The large performance swings from one epoch to the next reflects the random nature of augmentation differences between epochs. The variations would be less with a larger dataset or with a smaller learning rate.

\section{Results}
\label{sec:results}
\begin{figure}[t!]
\captionsetup{width=6cm}
    \centering
    \begin{minipage}[t]{0.5\textwidth}
    \vspace{0pt}
        \centering
        {%
        \setlength{\fboxrule}{1pt}%
        \includegraphics[width=0.95\textwidth]{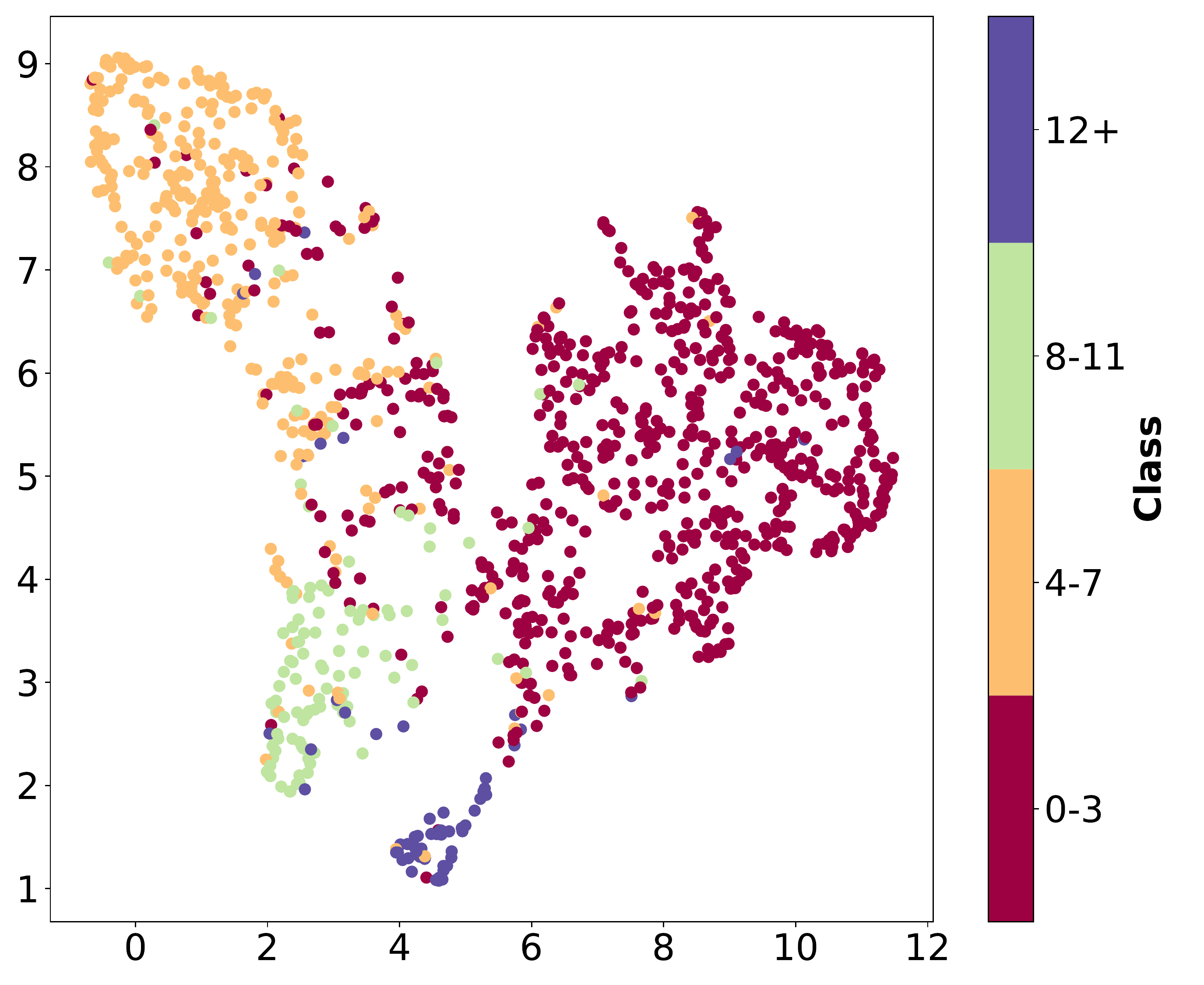}%
        }%
        \caption[UMAP Projection]{UMAP Projection of ResNeXt-50 model for fold-1} 
        \label{fig:umap}
    \end{minipage}\hfill
    \begin{minipage}[t]{0.5\textwidth}
    \vspace{0pt}
        \centering
        {%
        \setlength{\fboxrule}{1pt}%
        \includegraphics[width=0.95\textwidth]{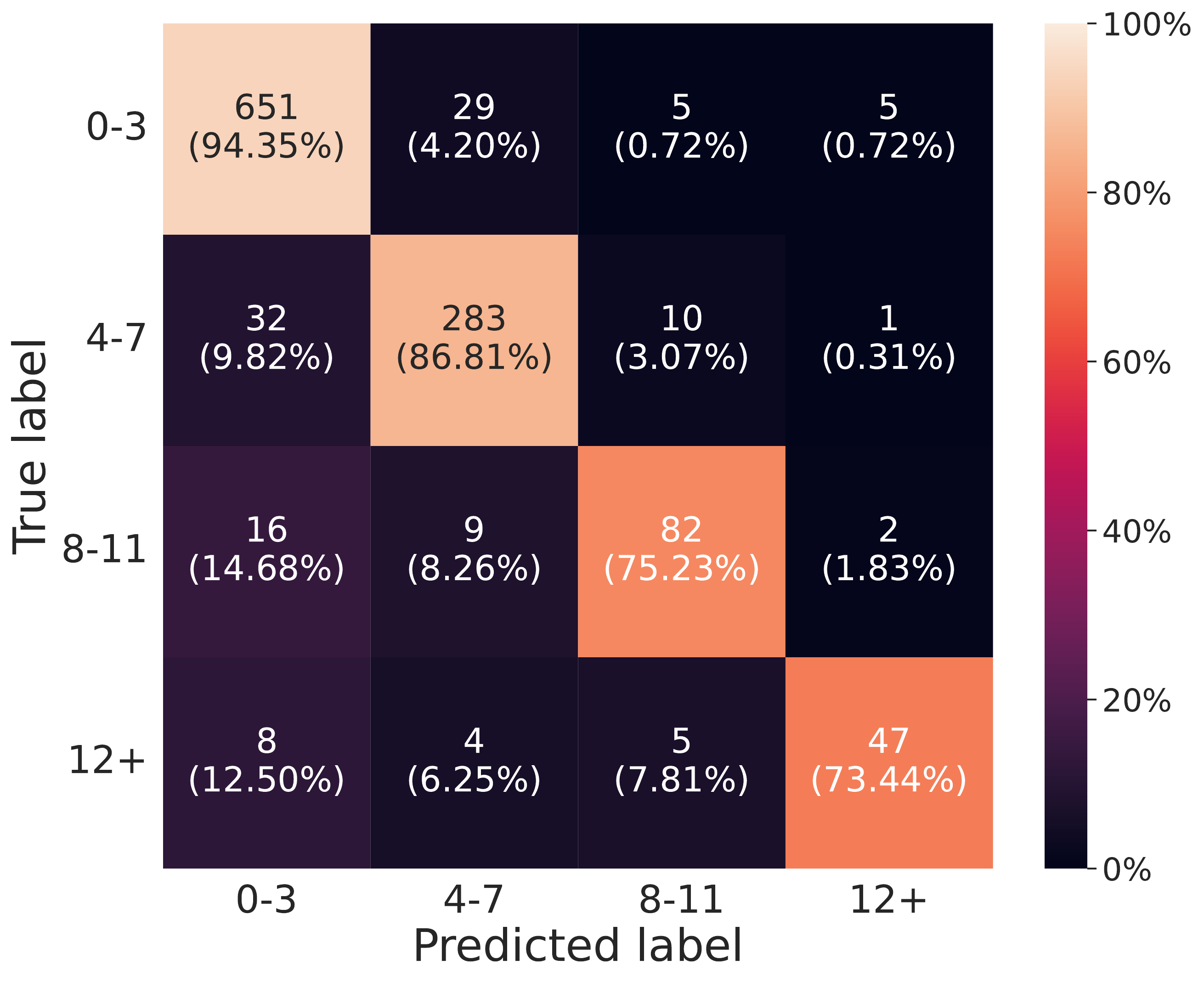}%
        }%
        \caption[Test Set Confusion Matrix]{ResNeXt-50 held out test set confusion matrix for fold-1}
        \label{fig:TestConfusion}
    \end{minipage}
\end{figure}

Uniform Manifold Approximation and Projection (UMAP) is a dimension reduction technique that is used for visualising general non-linear dimension reduction. The visualised results, (Figs. \ref{fig:umap} and \ref{fig:TestConfusion}), show a UMAP clustering plot and a confusion matrix plot, detailing how the best performing ResNeXt model has classified each image (88.99\% accuracy) from within the held-out test dataset, for one shuffle, using a seed of 0 for reproducibility. Class imbalance shows up within each class' performance and sees the highest class (12+) being most effected between dataset shuffles due to having the least samples. Overall metrics are reported for 5-fold cross validation with the best fold shown within Table \ref{tbl:fivefold}. The 5-Fold summary (Table. \ref{tbl:fivefoldsummary}) has a mean accuracy of 88\% with precision, recall and $F_{1}$-score metrics calculated using \emph{macro} averaging, $P_{Avg} = (P_{class 1} + P_{class 2} + P_{class 3} + P_{class 4}) / N_{classes}$.          

\begin{table}[t!]
\centering
\scriptsize
    {%
    \noindent\begin{tabular}{c|c|c|c|c|c}
        \textbf{Class} & \textbf{Precision} & \textbf{Recall} & \textbf{F1-Score} & \textbf{Accuracy \%} & \textbf{\# Images} \\
        \hline\hline
        0 - 3 & 0.92 & 0.94 & 0.93 & 94.35 & 690 \\
        4 - 7 & 0.87 & 0.87 & 0.87 & 86.81 & 326 \\
        8 - 11 & 0.80 & 0.75 & 0.78 & 75.23 & 109 \\
        12+ & 0.85 & 0.73 & 0.79 & 73.44 & 64 \\
        \hline\hline
        \textbf{Accuracy} & & & & 89.40 & 1189 \\
        \textbf{Macro Avg.} & 0.86 & 0.82 & 0.84 & 89.40 & 1189 \\
        \textbf{Weighted Avg.} & 0.89 & 0.89 & 0.89 & & 1189 \\
        \hline
    \end{tabular}
    }
\caption{Best performing fold by accuracy of ResNeXt-50 model for 5-Fold with seed 0}
\label{tbl:fivefold}
\end{table}%

\begin{table}[t!]
\captionsetup{width=9cm}
\centering
\scriptsize
    {%
    \noindent\begin{tabular}{c|c|c|c|c}
        \textbf{Fold} & \textbf{Precision} & \textbf{Recall} & \textbf{F1-Score} & \textbf{Accuracy} \\
        \hline\hline
        1 & 0.86 & 0.82 & 0.84 & 0.89\\
        2 & 0.85 & 0.81 & 0.83 & 0.88\\
        3 & 0.83 & 0.79 & 0.81 & 0.86\\
        4 & 0.85 & 0.78 & 0.81 & 0.87\\
        5 & 0.85 & 0.77 & 0.80 & 0.88 \\
        \hline
    \end{tabular}
    }
\caption{ 5-Fold held-out test set results of ResNeXt-50 model with seed 0. Precision, Recall and F1-Score are macro averages}
\label{tbl:fivefoldsummary}
\end{table}%

\begin{figure}[t!]
    \centering
    \captionsetup{width=0.5\textwidth}
        \includegraphics[width=0.6\textwidth]{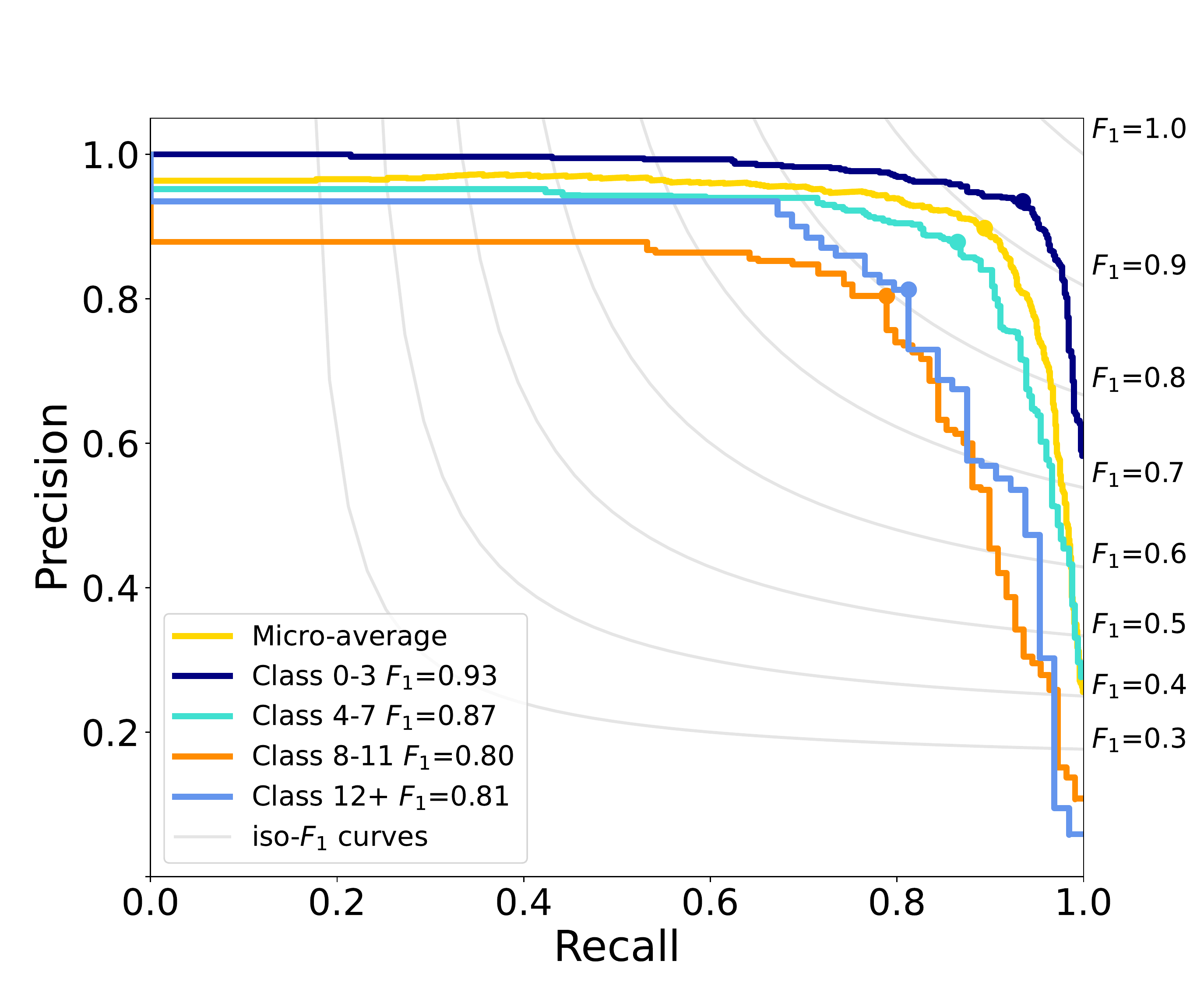}
        \caption[Precision-Recall Curves]{Precision Recall curves with interpolated precision class plots for fold-1 }
        \label{fig:KML-RoadFeatures1}
\end{figure}

\section{Discussion}
\label{sec:discussion}
While work continues on this project, it is not currently known how many classes can be distinguished from. The results show a 4 class system performs well, although less classes have shown better accuracy (well into the 90\% bracket) and a 5 class output also produces accuracy above 80\%. Class imbalance of verge samples provide a big challenge since over sampling minority classes, using augmentation, is not a good substitute for original images of verges with a higher number of positive indicator species. A larger dataset with more samples from the minority classes, rectifying class imbalance with real data, could possibly increase performance of all classes to $90\%+$ accuracy. On going work will interpolate between road features, Fig. \ref{fig:KML-RoadFeatures1}, giving more panorama locations to extract more higher scoring samples to alleviate class this imbalance. Different techniques addressing class imbalance such as a combination of under-sampling majority and over-sampling minority classes \citep{chawla_smote_2002} could also be employed. Understanding inter-class knowledge inside the CNN may also be a direction to explore \citep{wei_understanding_2015, NIPS2014_b86e8d03} as would be the effect that the overall image scene, sky and road, has upon classifying verges.  

Experiments (Sect. \ref{subsec:training}) would seem to backup comments \citep{DBLP:journals/corr/RazavianASC14} suggesting SOTA models can be taken \emph{off the shelf} and fine tuned, rather than hand crafting specific models for many applications, although perhaps with an added caveat of \emph{several models} should be tried before choosing. 

\section{Conclusion}
\label{sec:conclusion}
GSV imagery can be an effective data source for the remote surveying of roadside verges to a high probability that could identify verges that have little to no conservation potential. An automated pipeline enabling a SOTA CNN to learn both the fine differences between inter-class and broad differences within intra-class features is a possible remote replacement for large scale physical surveys that would normally take many hours spread over, potentially, years. Remote surveying or monitoring could be accomplished within hours for a fairly large area. With GPU / Computer technology constantly improving the situation year on year would afford either bigger areas to be surveyed within a usable time frame or surveys to be carried and results obtained within the same day. The implications of this study lie within the importance placed upon identification of LWSs for use within conservation management and environmental planning that local authorities are required to carry out. We Demonstrate that remote surveying using GSV as a data source for classification by a CNN is possible and would give authorities a cost effective way to manage and monitor their legal responsibilities whilst simultaneously aiding conservation efforts.    

\section{Acknowledgements}
\label{Acknowledgements}
This project would not have been possible without the Lincolnshire Wildlife Trust's \citep{LWT1} invaluable survey results from their Life on the Verge projects \citep{LotV1}. The thousands of hours work put in by 250 volunteers years ago are still being appreciated. 

\bibliographystyle{elsarticle-harv} 
\bibliography{cas-refs}





\end{document}